%% file: main.tex
\documentclass[11pt]{article}

\usepackage[numbers]{natbib}

\usepackage{wrapfig}
\usepackage{booktabs}       
\usepackage{amsfonts}       
\usepackage{amsmath}
\usepackage{graphicx}
\usepackage{amsthm}
\usepackage{amssymb}

\usepackage{multirow}

\usepackage[vlined,linesnumbered,ruled,resetcount]{algorithm2e}
\usepackage[colorlinks,linkcolor=magenta,filecolor=blue,citecolor=blue,urlcolor=blue]{hyperref}%
\usepackage[top=1in, left=1in, right=1in, bottom=1in]{geometry}
\usepackage{subfigure}
\usepackage{amssymb}
\usepackage{pifont}
\usepackage{mathtools}
\usepackage{stmaryrd}
\usepackage[T1]{fontenc}

\usepackage{xcolor}         % colors
\usepackage{pifont}

\definecolor{redx}{RGB}{180,0,0}
\definecolor{greenx}{RGB}{0,180,0}

\definecolor{redx}{RGB}{180,0,0}
\definecolor{greenx}{RGB}{0,180,0}

\usepackage{authblk}
\usepackage{enumitem}
\usepackage{algorithmic}

\theoremstyle{definition}

\usepackage{times}
\usepackage{latexsym}

\usepackage[T1]{fontenc}

\usepackage[utf8]{inputenc}

\usepackage{microtype}

\usepackage{inconsolata}

\usepackage{url}           
\usepackage{nicefrac}      
\usepackage{microtype}    
\usepackage{xcolor}        
\usepackage{listings}

\lstdefinelanguage{JSON}{
    basicstyle=\ttfamily,
    string=[s]{"}{"},
    comment=[l]{//},
    morecomment=[s]{/*}{*/},
    morestring=[b]',
    literate=
     *{0}{{{\color{black}0}}}{1}
      {1}{{{\color{black}1}}}{1}
      {2}{{{\color{black}2}}}{1}
      {3}{{{\color{black}3}}}{1}
      {4}{{{\color{black}4}}}{1}
      {5}{{{\color{black}5}}}{1}
      {6}{{{\color{black}6}}}{1}
      {7}{{{\color{black}7}}}{1}
      {8}{{{\color{black}8}}}{1}
      {9}{{{\color{black}9}}}{1}
}

\usepackage{tablefootnote}
\usepackage{amsmath}
\usepackage{enumerate}
\usepackage{tabularx}
\usepackage{booktabs}
\usepackage{esvect}
\usepackage{appendix}
\usepackage{import}
\usepackage{subcaption}
\usepackage{enumitem}
\usepackage{booktabs}
\usepackage{graphicx}

\usepackage{array}
\usepackage{booktabs}
\usepackage{multirow}
\usepackage{amssymb}
\usepackage{pifont}

\makeatletter
\newcommand*{\RN}[1]{\expandafter\@slowromancap\romannumeral #1@}
\makeatother

\makeatletter
\newcommand{\printfnsymbol}[1]{%
  \textsuperscript{\@fnsymbol{#1}}%
}
\makeatother

\title{Alopex: A Computational Framework for Enabling On-Device Function Calls with LLMs}

\author[1]{Yide Ran\thanks{Now with Stevens Institute of Technology}}
\author[1]{Zhaozhuo Xu\printfnsymbol{1}}
\author[1]{Yuhang Yao}
\author[1]{Zijian Hu}
\author[1]{Shanshan Han\thanks{With University of California, Irvine}}
\author[1]{Han Jin}
\author[1]{Alay Dilipbhai Shah}
\author[2]{Jipeng Zhang}
\author[1]{Dimitris Stripelis}
\author[3]{Tong Zhang}
\author[1]{Salman Avestimehr}
\author[1]{Chaoyang He}

\affil[1]{TensorOpera Inc.}

\affil[2]{Hong Kong University of Science and Technology}
\affil[3]{University of Illinois Urbana-Champaign}

\date{}
\begin{document}

\maketitle

\input{sections/abstract}

\input{sections/intro}

\input{sections/related}

\input{sections/method}

\input{sections/exp}

\input{sections/conclusion}

\bibliography{sections/ref}
\bibliographystyle{plainnat}
\clearpage

\appendix
\input{sections/appendix1}
\clearpage

\end{document}

%% file: sections/abstract.tex
\begin{abstract}
The rapid advancement of Large Language Models (LLMs) has led to their increased integration into mobile devices for personalized assistance, which enables LLMs to call external API functions to enhance their performance. However, challenges such as data scarcity, ineffective question formatting, and catastrophic forgetting hinder the development of on-device LLM agents.
To tackle these issues, we propose Alopex, a framework that enables precise on-device function calls using the Fox LLM. Alopex introduces a logic-based method for generating high-quality training data and a novel ``description-question-output'' format for fine-tuning, reducing risks of function information leakage. Additionally, a data mixing strategy is used to mitigate catastrophic forgetting, combining function call data with textbook datasets to enhance performance in various tasks. 
Experimental results show that Alopex improves function call accuracy and significantly reduces catastrophic forgetting, providing a robust solution for integrating function call capabilities into LLMs without manual intervention.

\end{abstract}

%% file: sections/intro.tex
\section{Introduction}

With the rapid advancement of Large Language Models (LLMs)~\cite{openai2024gpt4technicalreport, llama3modelcard, jiang2023mistral7b}, their integration into software applications has become increasingly widespread~\cite{Osika_gpt-engineer_2023, Significant_Gravitas_AutoGPT}. Researchers and engineers from both academia and industry are now focusing on developing LLM-powered agents~\cite{nakano2022webgptbrowserassistedquestionansweringhuman, weng2023agent} for mobile devices to provide personalized assistance to users. A key aspect of creating an on-device LLM agent is enabling LLMs to call external API functions~\cite{berkeley-function-calling-leaderboard, parisi2022talmtoolaugmentedlanguage} for enhanced performance. These API functions act as external tools, grounding the LLMs to generate more accurate and personalized outputs.

Despite the growing interest, several challenges hinder the development of on-device LLM agents with function call capabilities. We summarize these challenges from the perspective of data scarcity, question format, and catastrophic forgetting.

\noindent \textbf{Scarcity of LLM Function Call Demonstrations.} Training data that demonstrates successful LLM function calls is scarce. Since the concept of LLM agents has only recently gained researchers' attention, there is a shortage of examples showing how LLMs can effectively use external API functions on mobile devices to complete queries correctly. Consequently, current methods often rely on LLMs to generate synthetic function call demonstrations based on function descriptions~\cite{chen2024octopusv2ondevicelanguage}. However, this synthetic data, as seen in other domains~\cite{setlur2024rl}, raises concerns about correctness and information leakage~\cite{eldan2023s,aditya2024evaluating}. Therefore, a manual verification and regeneration process is necessary for these LLM-generated demonstrations, which impacts overall workflow efficiency.

\noindent \textbf{Effective Question Format Remains To Be Explored.} 
The optimal way to format queries to an LLM for successfully triggering the correct functions is still uncertain. This format needs to be consistent during both the training and inference phases to ensure aligned performance. One current approach~\cite{chen2024octopusv2ondevicelanguage} uses a ``question-call-description'' format, where 
the question, such as "Can I capture a high-resolution image from the front camera?", is placed first. Following the question, the function call command to initiate the function call procedure is placed next. Finally, the description of the triggered function is provided in the last section. However, this format presents two main issues, including 
\textit{i}) the function call command has limited prior in-context knowledge about the function descriptions, leading to potential inaccuracies; and \textit{ii}) if we fine-tune an LLM using this format, it will generate the function descriptions after the function call command during inference. Although manual intervention can stop this generation, it is vulnerable to malicious attacks. Therefore, we still need to identify an effective fine-tuning data format for  LLMs in terms of function calls.

\noindent \textbf{Significant Catastrophic Forgetting Happened in Function Call Fine-Tuning.} 
Existing LLM fine-tuning approaches for function calls may result in catastrophic forgetting~\cite{mccloskey1989catastrophic, ratcliff1990connectionist}, which significantly impairs
LLM performance on reasoning tasks. For instance, when evaluating fine-tuned LLMs with function calls on the MMLU dataset~\cite{hendrycks2020measuring}, a 16\% drop in accuracy is observed, making the fine-tuned LLMs perform like random guesses in high school multiple-choice questions. This study highlights the typical catastrophic forgetting phenomenon during LLM fine-tuning for function calls. In on-device LLM agents, it is crucial to maintain advanced LLM functionalities, such as commonsense and mathematical reasoning, alongside function calling. Therefore, it is imperative to develop fine-tuning strategies that prevent catastrophic forgetting while enabling LLMs for function calls.

To tackle these issues, we propose Alopex, a computational framework designed to enable on-device function calls using Fox LLM~\cite{foxllm} for precise, domain-specific responses. We summarize the contributions of Alopex below. 

\begin{itemize}[nosep,leftmargin=*]
\item We argue that function call demonstrations contain strong logic. Moreover, we propose a simple yet effective method to generate high-quality training demonstrations for LLMs to use API functions. Our study suggests that this logic-based demonstration generation approach leads to better fine-tuned LLMs for function calls.
\item Through our exploration, we present a new ``description-question-output'' data format for fine-tuning LLMs for function calls. We observe that this format performs better than the existing formats. Additionally, this format reduces the potential risks of leaking function information.
\item We introduce a data mixing approach to overcome catastrophic forgetting in fine-tuning LLMs for function calls. By combining our function call dataset with a textbook dataset, we demonstrate that the fine-tuned LLMs perform better in both function calls and other LLM evaluation benchmarks.
\end{itemize}

Moreover, we propose a series of system-level optimizations to integrate function call capabilities into LLMs without requiring manual verification and modification. Experimental results demonstrate that our framework achieves better accuracy in function calls compared to existing fine-tuned LLMs. Additionally, Alopex significantly reduces the catastrophic forgetting phenomenon observed in existing fine-tuned LLMs for function calls. Furthermore, our Alopex framework supports an automatic LLM adaptation pipeline encompassing data generation and LLM fine-tuning.

%% file: sections/related.tex
\section{Related Work}
\begin{figure*}[!ht]
  \centering
  \includegraphics[width=\textwidth]{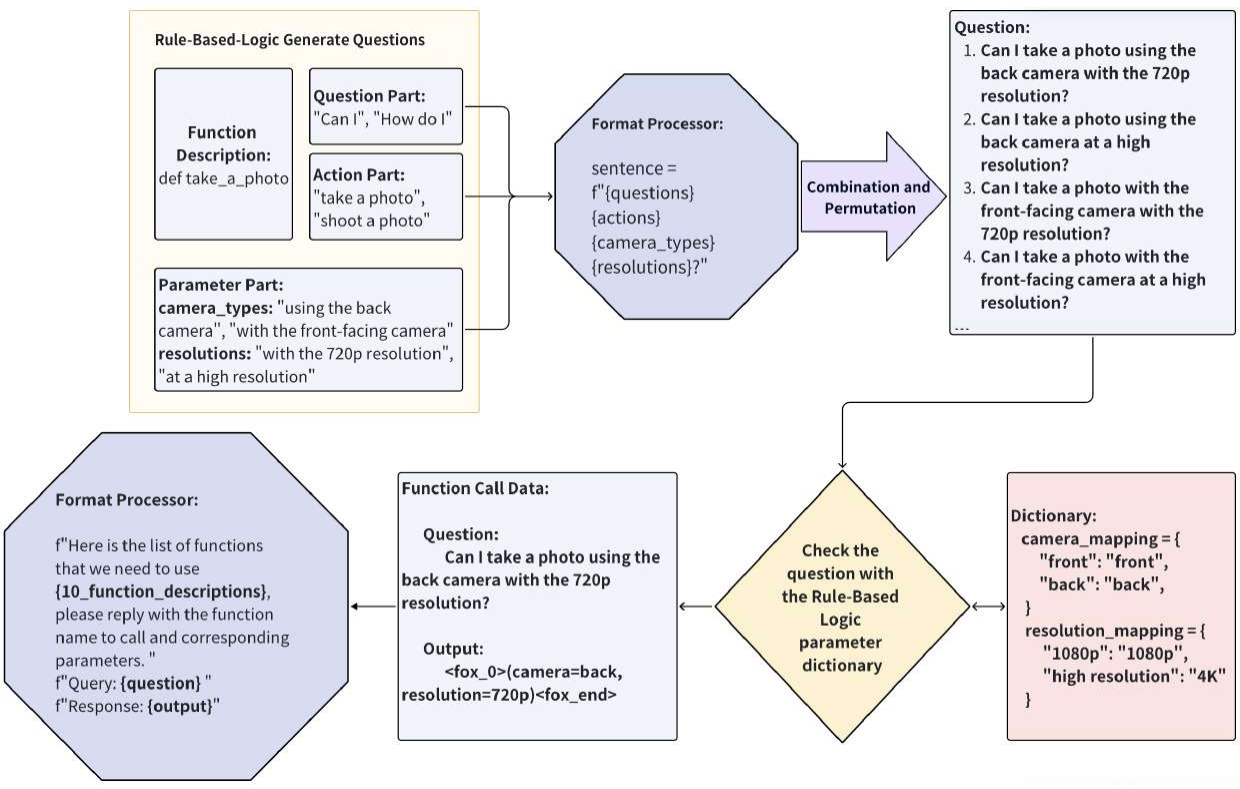}
  \caption{Rule-Based Logic Data Generation Work Flow, regarding the detail content of function description of $take\_a\_photo$, please refer to the content mentioned above in this study.}
  \label{fig:overflow}
\end{figure*}

We summarize the related works from the dataset generation, training schema, and benchmarks.

\noindent \textbf{Dataset Generation.}
Recent works have developed function call dataset generation pipeline. Octopus-v2~\cite{chen2024octopusv2ondevicelanguage} presents a methodology to use LLM to sample queries from open-source docstrings that are relevant to functions. Then, based on the sampled queries, LLM is used to generate outputs. Dialogue State Tracking (DST)  employs LLMs to generate dialogue
data to reduce dialogue collection and annotation costs~\cite{niu2024enhancingdialoguestatetracking}. APIGen~\cite{liu2024apigenautomatedpipelinegenerating} regenerates query data
for the APIs that have noisy and unusable descriptions, and design an automated pipeline to check and verify the Function-Calling Datasets. ToolAlpaca, MetaTool, Octo-planner and ``Self-Taught Reasoner'' (STR) also leverage the generative
capabilities of LLM to create comprehensive documentation
for each tool~\cite{huang2024metatoolbenchmarklargelanguage, tang2023toolalpacageneralizedtoollearning, chen2024octoplannerondevicelanguagemodel, zelikman2022starbootstrappingreasoningreasoning}. These

approaches all employ LLM for data generation. However, in practical implementation, considering the amount of data and the inference time of LLMs, generating data with LLMs is not highly efficient. Moreover, the results of data generation may contain errors, such as duplicate queries, incorrect function selection, incorrect parameter usage, and irrelevant outputs to the queries. To address the issue of data generation errors, the aforementioned works primarily rely on manual secondary checks or automated detection pipeline, which requires additional time costs for the inspection process. 

\noindent \textbf{Training Schema.}
Previous works, such as ToolAlpaca~\cite{tang2023toolalpacageneralizedtoollearning} and Octopus-v2, rely on supervised fine-tuning to enable LLMs with function call capabilities. However, this approach neglects the issue of catastrophic forgetting caused by fine-tuning negative impact on LLMs. Enabling LLM for function calls share spirits with the LLM alignment techniques~\cite{lin2023speciality}. 

\noindent \textbf{Benchmarks.}
For this study, we adopt the Berkeley Function-Calling Leaderboard~\cite{berkeley-function-calling-leaderboard} as our evaluation metric for single function call accuracy, and utilize the open-source library lm-auto-evaluation~\cite{eval-harness} to evaluate the LLMs performance on metrics MMLU~\cite{hendrycks2020measuring}, GSM8K~\cite{cobbe2021trainingverifierssolvemath}, ARC~\cite{clark2018think}, TruthfulQA~\cite{lin2022truthfulqa}, HellaSwag~\cite{zellers2019hellaswag}, and Winogrande~\cite{DBLP:journals/corr/abs-1907-10641}.

%% file: sections/method.tex
\section{Alopex Framework}
We overview our framework in Figure 1. The framework contains three major components: \textit{i}) function call demonstration generation, \textit{ii}) formatting function call demonstrations for LLM fine-tuning, and \textit{iii}) overcoming catastrophic forgetting.

\subsection{Generating Function Call Demonstrations Using Rule-Based Logic}

In this study, we employed the Rule-Based Logic approach to generate questions and outputs for the dataset. Based on the function description of each API, we categorized common user questions into two styles: \textit{requests} and \textit{commands}. Requests can be further divided into a question part, an action part, and a parameter part. 
Commands, on the other hand, only include the action and the parameter parts.

Let's consider the $take\_a\_photo$ API as an example. 
For more function descriptions, please refer to \S\ref{sec:app_func_describe} in the appendix.
Based on the given function description, we can generate the detailed elements that constitute a question, following the division into the question part, the action part, and the parameter part. For the content of the three-part elements, please refer to \S\ref{sec:app_func_describe} in the appendix.

By arranging the elements in the order of question, action, and parameter, we can instantly generate questions related to function $take\_a\_photo$. As long as the number of queries we want to generate is less than the maximum combinations of questions, actions, and parameters, we don't have to worry about generating duplicate questions. For generating the outputs, we can establish a mapping table between parameters and their corresponding values. Based on the parameter content appearing in the question, we can retrieve the corresponding values from the mapping table to generate the value part of the output.
By following the logic to generate data, we can make sure that we won't generate incorrect function parameters and function names. Simultaneously, we can greatly improve efficiency as well. We save a huge amount of time on LLM inference and eliminate the need for manual data checking and validation and LLM regeneration in the later stages.

\subsection{Formatting Function Call Demonstrations for LLM Fine-Tuning}
\begin{figure*}[!ht]
  \centering
  \includegraphics[width=\textwidth]{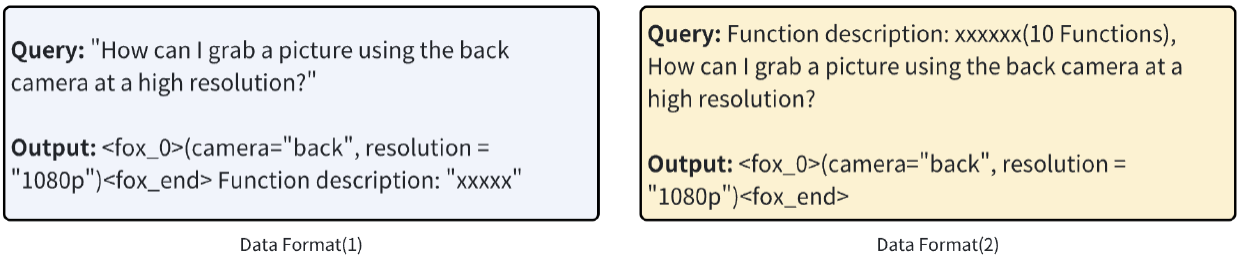}
  \caption{Data Format (1) is based on the format speculated to be used in Octopus-v2, while Data Format (2) is created using our approach.}
  \label{fig:data-format}
  \vspace{-.3cm}
\end{figure*}

There is still an issue with the dataset's structure regarding the position of the function description. Although Octopus-v2~\cite{chen2024octopusv2ondevicelanguage} has not open-source their dataset, based on the LLM inputs and outputs, we speculate that its dataset structure follows the format of  Data Format(1) in Figure~\ref{fig:data-format}, called as 
$\mathit{DF}_1$.

Meanwhile, 
according to the data training format provided by gorilla-llm/gorilla-openfunctions-v1~\cite{berkeley-function-calling-leaderboard}, they employ Data Format(2) in Figure~\ref{fig:data-format}, \textit{i}.\textit{e}., $\mathit{DF}_2$.

So, which one is better? Before we proceed with the next step, we need to introduce a metric called Out-of-Logic Function Call Accuracy (OOFC). Since we design the dataset generation rules manually, it is inevitable that cases will deviate from the pre-defined rules. For example, the existing designed rules generate the question part that contains phrases such as ``Can I'', 
``How do I'', ``How can I'', ``Is it possible to'', and ``I want to''. Any phrases like ``What's the approach to'', ``Is it achievable to'', ``I wish to'', ``Could you aid me'', or ``Would you assist me'' are considered as ``out-of-logic rules''. Therefore, the quality of the LLM fine-tuning data and the function call data can be evaluated by the performance of OOFC.
After conducting extensive experiments, we have discovered that the data structure follows $\mathit{DF}_2$ not only achieves \textbf{99\%} accuracy in function call accuracy when following logic but also performs better than $\mathit{DF}_1$ in out-of-logic scenarios. Therefore, we have chosen to use the function description and the question as the combined input.

\subsection{Overcoming Catastrophic Forgetting in LLM Fine-Tuning for Function Calls}

Previous studies have shown that after fine-tuning and continuing learning, LLMs can successfully perform downstream tasks. 
However, they may forget their previous knowledge and perform poorly on previous tasks. This
phenomenon is called catastrophic forgetting. One common approach to address catastrophic forgetting is to mix the history pretraining data with the new data to enable LLMs to recall previous knowledge~\cite{luo2024empiricalstudycatastrophicforgetting,shi2024continuallearninglargelanguage,huang2024mitigatingcatastrophicforgettinglarge}. 
We have also observed the phenomenon of catastrophic forgetting after fine-tuning  
the function call task. While the LLMs perform well on the function call task, there is
a significant decrease in metrics MMLU~\cite{hendrycks2020measuring}, GSM8K~\cite{cobbe2021trainingverifierssolvemath}, Arc~\cite{clark2018think}, HellaSwag~\cite{zellers2019hellaswag}, Winogrande~\cite{DBLP:journals/corr/abs-1907-10641}, and TruthfulQA~\cite{lin2022truthfulqa} compared with the pretraining LLM. 

Although mixing the pretraining dataset alleviates 
catastrophic forgetting, 
the pretraining data for LLMs is not publicly accessible. Therefore, the approach of mixing the pretraining dataset is not feasible.  
Empirically, textbook datasets are typically used for pretraining LLMs. Although we do not have access to the exact textbook datasets used for LLM pretraining, is it possible for us to use opensource textbook datasets for mixed dataset finetuning to achieve the same goal of mitigating catastrophic forgetting? We conducted experiments by mixing the function call dataset with the opensource dataset nampdn-ai/tiny-textbooks~\cite{nam_pham_2023} with ratio 1:1 and observed that the phenomenon of catastrophic forgetting can be mitigated, some metrics are even better than before. These experiments were performed on models google/gemma-2b, Qwen/Qwen1.5-1.8B, stabilityai/stablelm-2-1.6b, tensoropera/Fox-1-1.6B.

%% file: sections/exp.tex
\begin{table*}[!h]
\centering
\footnotesize
\caption{Data Generation Comparision Follow The Logic} 

\resizebox{\linewidth}{!}{
\begin{tabular}{lcccc}
\toprule[1.5pt]
Data Generation\textbackslash{}Model &  stabilityai/stablelm-2-1\_6b & Gemma2B  & Qwen1.5-1.8B & tensoropera/Fox-1-1.6B \\
\midrule[1.5pt]
Question:Output + func describe &  0.996 &  0.9945  &  0.9954 &  0.989  \\
\midrule[0.5pt]
LLM Generation (Octopus) &  0.705 & 0.719  &  0.697   & 0.732 \\
\midrule[0.5pt]
Function describtion + question & 0.999  & 0.997    & 0.997    &  0.997 \\
\bottomrule[1.5pt] 
\end{tabular}
}
\label{tab:data}
\centering
\footnotesize
\caption{Data Generation Comparison Out Of Logic} 
\vspace{0.0cm}

\resizebox{\linewidth}{!}{
\begin{tabular}{lcccc}
\toprule[1.5pt]
Data Generation\textbackslash{}Model &  stabilityai/stablelm-2-1\_6b & Gemma2B  & Qwen1.5-1.8B & tensoropera/Fox-1-1.6B \\
\midrule[1.5pt]
Question:Output + func describe &  0.92 &  0.967  &  0.971 &  0.964  \\
\midrule[0.5pt]
LLM Generation (Octopus) &  0.659 & 0.67  &  0.672   & 0.6709 \\
\midrule[0.5pt]
Function describtion+question & 0.974  & 0.984    & 0.972    &  0.975 \\
\bottomrule[1.5pt] 
\end{tabular}
}
\label{tab:data2}
\end{table*}
\section{Experiment}
In this section, we want to validate the effectiveness of the Alopex framework. We would like to answer the following questions:

\begin{itemize}[nosep,leftmargin=*]
    \item \textbf{Q1:} Without conducting an examination on the generated dataset, how much better is the Rule-Based Logic method compared with the LLM-based method? 

    \item \textbf{Q2:} Why does the data format $\mathit{DF}_1$ perform better than $\mathit{DF}_2$?

    \item \textbf{Q3:} How effective is the approach of mixing datasets in a 1:1 ratio in mitigating catastrophic forgetting?
\end{itemize}

\subsection{Experiment Settings}
This study focuses on single function calls. We utilized three constructed datasets, and
conducted fine-tuning experiments on the following models: google/gemma-2b~\cite{gemmateam2024gemmaopenmodelsbased}, stabilityai/stablelm-2-1\_6b~\cite{bellagente2024stable}, Qwen1.5 - 1.8B~\cite{qwen}, and tensoropera/Fox-1-1.6B~\cite{foxllm}. 
Due to the relatively small number of parameters 
of these LLMs, we directly fine-tuned the full models. The training parameters are provided in \S\ref{sec:hyper_parameters} in the appendix.

\subsubsection{Dataset}
We referenced 10 Android APIs open-sourced from Octopus-v2~\cite{chen2024octopusv2ondevicelanguage} to construct the function call datasets; see \S\ref{sec:app_func_describe} in the appendix for the descriptions of the 10 functions. 
We generated the following three datasets for our research questions.

\begin{itemize}[nosep, leftmargin=*]
    \item We generated 1,000 data points for each API using a Logic-Rule Based method. The format of the dataset follows $\mathit{DF}_1$. The dataset was then split into a training dataset containing 200 data items and a test dataset containing 800 data items.
    \item We employed the GPT3.5 API~\cite{openai_api}
    to generate the dataset, using the  Octopus-v2 approach~\cite{chen2024octopusv2ondevicelanguage}. Since we did not have access to the open-source docstrings mentioned in Octopus-v2, we directly utilized the queries generated in the first part of our dataset generation as the docstrings. 
    We mixed and shuffled the question dataset generated by the Rule-Based Logic method for the 10 functions. Then, using the GPT-3.5 API, we filtered the queries based on each function's description. Subsequently, we used the GPT-3.5 API to generate the output that contains the function name and the function parameters. Additionally, we appended each function's description after the output. We used the test dataset generated from previous rule-based logic to evaluate the performance of LLMs trained using the LLM-generated data.    

    \item 
    We generate a dataset with the same generation process and a different data format  $\mathit{DF}_2$. Note that we 
    utilized the 10 function descriptions and the user's questions as inputs, instead of only using the user's questions as inputs, as the model needs to select the most appropriate function from the 10 function descriptions based on the user's questions to assist the user.
    
\end{itemize}

\input{sections/table_exp3}

\noindent We employed these three of datasets to fine-tune Qwen1.5-1.8B, Gemma2B, stabilityai/stablelm-2-1.6b, and tensoropera/Fox-1-1.6B.

\subsubsection{Settings}

\noindent\textbf{Testbed.} 
We conducted our experiments using 2 GPUs NVIDIA H100 80GB HBM3.

\noindent\textbf{Evaluation Benchmarks and Metrics.}
We utilized the algorithm from the Berkeley Function Call Leaderboard~\cite{berkeley-function-calling-leaderboard} to calculate the function call accuracy. Additionally, we employed the LM Evaluation Harness~\cite{eval-harness} to measure metrics such as MMLU, GSM8K, Arc, HellaSwag, Winogrande, TruthfulQA, and their average. Additionally, we introduced a new metric called Function Call LLM Average Accruacy(FCLAA)which is the sum of average function call accuracy in Rule domain and Out of rule domain, and the average accuracy of MMLU, GSM8K, Arc, HellaSwag, Winogrande, TruthfulQA into our evaluation. Experiments revealed that the function call data generated by GPT-3.5 alone has many errors. It can only achieve an accuracy of 70\% which required subsequent validation, modification and re-generation. The Rule-Based Logic method is efficient and generates high-quality function call data. LLMs using Rule-Based Logic function call data finetuning can achieve 99\% accuracy. Since our output does not directly use the function name but instead utilizes special tokens inspired by Octopus-v2~\cite{chen2024octopusv2ondevicelanguage} to reduce the probability of spelling errors in predicting function names, we used 10 shots to provide prompts for the expected output when testing the ability of pre-trained LLMs to perform function calls. The experiment revealed that, apart from gemma-2b, none of the other three LLMs were able to perform function calls, with gemma-2b exhibiting an accuracy of 48.9\% for function calls.

\subsection{Results}
\noindent 
\textbf{Exp1: Comparisons between the Rule-Based Logic method and the LLM-based method. }
Based on the experimental results from Table~\ref{tab:data} and Table~\ref{tab:data2}, we found that while using GPT3.5 to generate data for function calls allows for automated selection of queries that are suitable for each function and automated batch generation of function call outputs, it still results in a significant number of errors, which affects the effectiveness of the fine-tuned LLMs.
Consequently, careful checking, validation and secondary generation are required after generating the datasets.

 \noindent \textbf{Exp 2: Comparisons between $\mathit{DF}_1$ and $\mathit{DF}_2$.}
Based on the experimental results from Table~\ref{tab:data} and Table~\ref{tab:data2},
we observed that $\mathit{DF}_2$ was more likely to get better function call accuracy, especially in the out-of-logic case.

\noindent \textbf{Exp 3: Evaluation of the effectiveness of mixing datasets.}
\noindent Based on the results in Table~\ref{tab:hyperparameter}, using open-source textbook datasets helps recover LLMs' previous performance. Although there are fluctuations in metrics, the average accuracy of all metrics shows an upward trend, except for Fox, which is more robust and exhibits stable results and achieves the highest average accuracy.

%% file: sections/table_exp3.tex
\begin{table*}[!h]
\centering
\caption{Details of Performance Metrics.} 
\vspace{0.0cm}

\resizebox{\linewidth}{!}{
\begin{tabular}{lccccccccccc}
\toprule[1.5pt]
Model &  Training Strategy & Func Call ACC & OOFC &   MMLU    & GSM8K  &  Arc &  HellaSwag & Winogrande & TruthfulQA & Average Acc \\
\midrule[1.5pt]
\textbf{Octopus-v2} & NA & 0.99048  & 0.990   & 0.2574  & 0.0023  &  0.3302 & 0.5000 &  0.5722 & 0.4149 & 0.775  \\\hline
\multirow{5}{4.3cm}{\textbf{stabilityai/stablelm-2-1.6b}} & NA &  0.001  & 0.001   & 0.3916 & 0.1766  &  0.4326 & 0.7035 &  0.6535 & 0.3877 & 0.152  \\
 & Output + Describe & 0.996 &  0.92 & 0.2603  & 0.0008   &  0.3797 & 0.6077 &  0.6093 & 0.4100 & 0.764  \\
 & Mix(1:1) Output+Describe & 0.99  & 0.943  & 0.3655 & 0.1221  &  0.4027  & 0.7020 & 0.6085 & 0.3688 & 0.7871  \\
 & 10 Func Describe + question & \textbf{0.999} &  0.93 & 0.3304  & 0.0212   &  0.4002 & 0.6291 &  0.6298 & 0.3212 & 0.769  \\
 & Mix(1:1) Func Describe + question & 0.994 & \textbf{0.971} & 0.3477 & 0.1205 & 0.4002 & 0.6593 & 0.6148 & 0.3778 & \textbf{0.795} \\\hline
\multirow{5}{4.3cm}{\textbf{Gemma2B}} & NA & 0.489 & 0.449 & 0.4174 &  0.1812 & 0.4923 & 0.7154  & 0.6575  & 0.3308 & 0.4679   \\
& Output + Describe & 0.9945 & 0.967 & 0.2985 & 0.0129 & 0.4352 & 0.6314 & 0.6212 & 0.3892 & 0.786 \\
& Mix(1:1) Output+Describe & 0.9945 & 0.965 & 0.3600 & 0.1236 & 0.4701 & 0.6811 & 0.6346 & 0.3745 & 0.80 \\
& 10 Func Describe + question & 0.997 & 0.984
 & 0.2892 & 0.0334 & 0.4121 & 0.6319 & 0.6338 & 0.3607 & 0.7915 \\
 & Mix(1:1) Func Describe + question & \textbf{0.999} & \textbf{0.986} & 0.3608 & 0.1440 & 0.4616 & 0.6420 & 0.6346 & 0.3812 & \textbf{0.807} \\\hline
 \multirow{5}{4.3cm}{\textbf{Qwen1.5 - 1.8B}} & NA & 0.004 & 0.001 & 0.4710 & 0.3366 & 0.3686 & 0.6173 & 0.6148 & 0.3937 & 0.157 \\
 & Output + Describe & 0.9954 & 0.971 & 0.3685 & 0.0349 & 0.3464 & 0.5507 & 0.5912 & 0.3644 & 0.7808 \\
 & Mix(1:1) Output+Describe & 0.981 & 0.929 & 0.4685 & 0.3275 & 0.3754 & 0.6313 & 0.6283 & 0.3927 & 0.793 \\
 & 10 Func Describe + question & 0.997& \textbf{0.972} & 0.4467 & 0.2790 & 0.3976 & 0.5812 & 0.6022 & 0.3864 & 0.805 \\
 & Mix(1:1) Func Describe + question & \textbf{0.998} & 0.961 & 0.4615 & 0.3048 & 0.3848 & 0.6016 & 0.6038 & 0.4041 & \textbf{0.806} \\\hline
\multirow{5}{4.3cm}{\textbf{tensoropera/Fox-1-1.6B}} & NA & 0.102 & 0.095 & 0.4303 & 0.3654 & 0.4087 & 0.6273 & 0.6062 & 0.3866 & 0.223 \\
& Output + Describe & 0.989 & 0.964 & 0.4196 & 0.2858 & 0.3933 & 0.6197 & 0.5998 & 0.3879 & 0.801 \\
& Mix(1:1) Output+Describe & 0.990 & 0.966 & 0.4090 & 0.3374 & 0.4215 & 0.6476 & 0.6188 & 0.4046 & 0.8097 \\
& 10 Func Describe + question & 0.997 & 0.975 & 0.4193 & 0.3245 & 0.4010 & 0.6182 & 0.5951 & 0.4246 & \textbf{0.8119} \\
& Mix(1:1) Func Describe + question & \textbf{0.999} & \textbf{0.976} & 0.4065 & 0.2851 & 0.3763 & 0.6038 & 0.5856 & 0.4047
 & 0.806 \\
\bottomrule[1.5pt] 
\end{tabular}
}
\vspace{0.0cm}
\label{tab:hyperparameter}
\end{table*}

%% file: sections/conclusion.tex
\section{Conclusion}

Alopex efficiently generates high-quality function call datasets without the need for data validation and regeneration. It also alleviates the catastrophic forgetting phenomenon caused by the function call task in LLMs. Experimental results show that, among the four small LLMs, Fox performs the highest level of robustness and average accuracy.

%% file: sections/appendix1.tex
\section*{Appendix}
\section{Function Description}\label{sec:app_func_describe}

We present the question part, action part, and parameter part we designed for the Rule-Based Logic for the $take\_a\_photo$ API. We also show the 
parameter Dictionary Mapping for the $take\_a\_photo$ API.

\noindent To better compare with Octopus-v2~\cite{chen2024octopusv2ondevicelanguage}, the functions and their descriptions we are using here are from their study.

\section{Experiment Parameters}\label{sec:hyper_parameters}
In the experiment, we used the following configuration parameters:

\begin{itemize}[nosep,leftmargin=*]
    \item \textbf{Learning Rate:} 6e-5, 5e-5, 4e-5, 3e-5, 2e-5, 4e-4, 3e-4, 5e-4
    
    \item \textbf{Batch Size:} 1, 2
    
    \item \textbf{Epochs:} [3,8]
    
    \item \textbf{Warmup Steps:} 5
    
    \item \textbf{Max sequence length:} 2048
    
    \item \textbf{LR scheduler type:} linear
\end{itemize}

\noindent We conducted 10 experiments for each model using this set of hyperparameters. The best result from each experiment was selected for the report.

\begin{lstlisting}[language=JSON, basicstyle=\ttfamily\small, breaklines=true, columns=flexible, showstringspaces=false]
"actions" : [
    "take a photo", "snap a picture", "capture an image", "shoot a photo", "get a snapshot",
    "record a photo", "grab a picture", "click a photo", "take a selfie"
],
"camera_types" : [
    "using the back camera", "with the rear camera", "using the front camera",
    "with the front-facing camera", "on the rear camera", "on the front camera", ""
],
"resolutions" : [
    "with the 720p resolution", "with the 1080p resolution", "with the 4K resolution", "", "at a high resolution", "at a clear resolution", "at a relative low resolution"
],
"questions" : [
    "Can I", "How do I", "How can I", "Is it possible to", "What's the process for",
    "Is there a way to", "What's the easiest way to", "", "I want to", "Please help me", "Could you help me"
]
\end{lstlisting}

\begin{lstlisting}[language=JSON, basicstyle=\ttfamily\small, breaklines=true, columns=flexible, showstringspaces=false]
camera_mapping = {
    "front": "front",
    "back": "back",
    "rear": "back"
}
resolution_mapping = {
    "720p": "720p",
    "relative low resolution": "720p",
    "1080p": "1080p",
    "clear resolution": "1080p",
    "4K": "4K",
    "high resolution": "4K"
}
\end{lstlisting}

%% file: main.bbl
\begin{thebibliography}{38}
\providecommand{\natexlab}[1]{#1}
\providecommand{\url}[1]{\texttt{#1}}
\expandafter\ifx\csname urlstyle\endcsname\relax
  \providecommand{\doi}[1]{doi: #1}\else
  \providecommand{\doi}{doi: \begingroup \urlstyle{rm}\Url}\fi

\bibitem[Aditya et~al.(2024)Aditya, Chawla, Dhingra, Rai, Sood, Singh, Wase, Bahga, and Madisetti]{aditya2024evaluating}
Harshvardhan Aditya, Siddansh Chawla, Gunika Dhingra, Parijat Rai, Saumil Sood, Tanmay Singh, Zeba~Mohsin Wase, Arshdeep Bahga, and Vijay~K Madisetti.
\newblock Evaluating privacy leakage and memorization attacks on large language models (llms) in generative ai applications.
\newblock \emph{Journal of Software Engineering and Applications}, 17\penalty0 (5):\penalty0 421--447, 2024.

\bibitem[AI@Meta(2024)]{llama3modelcard}
AI@Meta.
\newblock Llama 3 model card.
\newblock 2024.
\newblock URL \url{https://github.com/meta-llama/llama3/blob/main/MODEL_CARD.md}.

\bibitem[Bai et~al.(2023)Bai, Bai, Chu, Cui, Dang, Deng, Fan, Ge, Han, Huang, Hui, Ji, Li, Lin, Lin, Liu, Liu, Lu, Lu, Ma, Men, Ren, Ren, Tan, Tan, Tu, Wang, Wang, Wang, Wu, Xu, Xu, Yang, Yang, Yang, Yang, Yao, Yu, Yuan, Yuan, Zhang, Zhang, Zhang, Zhang, Zhou, Zhou, Zhou, and Zhu]{qwen}
Jinze Bai, Shuai Bai, Yunfei Chu, Zeyu Cui, Kai Dang, Xiaodong Deng, Yang Fan, Wenbin Ge, Yu~Han, Fei Huang, Binyuan Hui, Luo Ji, Mei Li, Junyang Lin, Runji Lin, Dayiheng Liu, Gao Liu, Chengqiang Lu, Keming Lu, Jianxin Ma, Rui Men, Xingzhang Ren, Xuancheng Ren, Chuanqi Tan, Sinan Tan, Jianhong Tu, Peng Wang, Shijie Wang, Wei Wang, Shengguang Wu, Benfeng Xu, Jin Xu, An~Yang, Hao Yang, Jian Yang, Shusheng Yang, Yang Yao, Bowen Yu, Hongyi Yuan, Zheng Yuan, Jianwei Zhang, Xingxuan Zhang, Yichang Zhang, Zhenru Zhang, Chang Zhou, Jingren Zhou, Xiaohuan Zhou, and Tianhang Zhu.
\newblock Qwen technical report.
\newblock \emph{arXiv preprint arXiv:2309.16609}, 2023.

\bibitem[Bellagente et~al.(2024)Bellagente, Tow, Mahan, Phung, Zhuravinskyi, Adithyan, Baicoianu, Brooks, Cooper, Datta, et~al.]{bellagente2024stable}
Marco Bellagente, Jonathan Tow, Dakota Mahan, Duy Phung, Maksym Zhuravinskyi, Reshinth Adithyan, James Baicoianu, Ben Brooks, Nathan Cooper, Ashish Datta, et~al.
\newblock Stable lm 2 1.6 b technical report.
\newblock \emph{arXiv preprint arXiv:2402.17834}, 2024.

\bibitem[Chen and Li(2024)]{chen2024octopusv2ondevicelanguage}
Wei Chen and Zhiyuan Li.
\newblock Octopus v2: On-device language model for super agent, 2024.
\newblock URL \url{https://arxiv.org/abs/2404.01744}.

\bibitem[Chen et~al.(2024)Chen, Li, Guo, and Shen]{chen2024octoplannerondevicelanguagemodel}
Wei Chen, Zhiyuan Li, Zhen Guo, and Yikang Shen.
\newblock Octo-planner: On-device language model for planner-action agents, 2024.
\newblock URL \url{https://arxiv.org/abs/2406.18082}.

\bibitem[Clark et~al.(2018)Clark, Cowhey, Etzioni, Khot, Sabharwal, Schoenick, and Tafjord]{clark2018think}
Peter Clark, Isaac Cowhey, Oren Etzioni, Tushar Khot, Ashish Sabharwal, Carissa Schoenick, and Oyvind Tafjord.
\newblock Think you have solved question answering? try arc, the ai2 reasoning challenge, 2018.

\bibitem[Cobbe et~al.(2021)Cobbe, Kosaraju, Bavarian, Chen, Jun, Kaiser, Plappert, Tworek, Hilton, Nakano, Hesse, and Schulman]{cobbe2021trainingverifierssolvemath}
Karl Cobbe, Vineet Kosaraju, Mohammad Bavarian, Mark Chen, Heewoo Jun, Lukasz Kaiser, Matthias Plappert, Jerry Tworek, Jacob Hilton, Reiichiro Nakano, Christopher Hesse, and John Schulman.
\newblock Training verifiers to solve math word problems, 2021.
\newblock URL \url{https://arxiv.org/abs/2110.14168}.

\bibitem[Eldan and Russinovich(2023)]{eldan2023s}
Ronen Eldan and Mark Russinovich.
\newblock Who's harry potter? approximate unlearning in llms.
\newblock \emph{arXiv preprint arXiv:2310.02238}, 2023.

\bibitem[Gao et~al.(2023)Gao, Tow, Abbasi, Biderman, Black, DiPofi, Foster, Golding, Hsu, Le~Noac'h, Li, McDonell, Muennighoff, Ociepa, Phang, Reynolds, Schoelkopf, Skowron, Sutawika, Tang, Thite, Wang, Wang, and Zou]{eval-harness}
Leo Gao, Jonathan Tow, Baber Abbasi, Stella Biderman, Sid Black, Anthony DiPofi, Charles Foster, Laurence Golding, Jeffrey Hsu, Alain Le~Noac'h, Haonan Li, Kyle McDonell, Niklas Muennighoff, Chris Ociepa, Jason Phang, Laria Reynolds, Hailey Schoelkopf, Aviya Skowron, Lintang Sutawika, Eric Tang, Anish Thite, Ben Wang, Kevin Wang, and Andy Zou.
\newblock A framework for few-shot language model evaluation, 12 2023.
\newblock URL \url{https://zenodo.org/records/10256836}.

\bibitem[Hendrycks et~al.(2020)Hendrycks, Burns, Basart, Zou, Mazeika, Song, and Steinhardt]{hendrycks2020measuring}
Dan Hendrycks, Collin Burns, Steven Basart, Andy Zou, Mantas Mazeika, Dawn Song, and Jacob Steinhardt.
\newblock Measuring massive multitask language understanding.
\newblock \emph{arXiv preprint arXiv:2009.03300}, 2020.

\bibitem[Huang et~al.(2024{\natexlab{a}})Huang, Cui, Wang, Yang, Liao, Song, Yao, and Su]{huang2024mitigatingcatastrophicforgettinglarge}
Jianheng Huang, Leyang Cui, Ante Wang, Chengyi Yang, Xinting Liao, Linfeng Song, Junfeng Yao, and Jinsong Su.
\newblock Mitigating catastrophic forgetting in large language models with self-synthesized rehearsal, 2024{\natexlab{a}}.
\newblock URL \url{https://arxiv.org/abs/2403.01244}.

\bibitem[Huang et~al.(2024{\natexlab{b}})Huang, Shi, Li, Fan, Wu, Zhang, Liu, Zhou, Wan, Gong, and Sun]{huang2024metatoolbenchmarklargelanguage}
Yue Huang, Jiawen Shi, Yuan Li, Chenrui Fan, Siyuan Wu, Qihui Zhang, Yixin Liu, Pan Zhou, Yao Wan, Neil~Zhenqiang Gong, and Lichao Sun.
\newblock Metatool benchmark for large language models: Deciding whether to use tools and which to use, 2024{\natexlab{b}}.
\newblock URL \url{https://arxiv.org/abs/2310.03128}.

\bibitem[Jiang et~al.(2023)Jiang, Sablayrolles, Mensch, Bamford, Chaplot, de~las Casas, Bressand, Lengyel, Lample, Saulnier, Lavaud, Lachaux, Stock, Scao, Lavril, Wang, Lacroix, and Sayed]{jiang2023mistral7b}
Albert~Q. Jiang, Alexandre Sablayrolles, Arthur Mensch, Chris Bamford, Devendra~Singh Chaplot, Diego de~las Casas, Florian Bressand, Gianna Lengyel, Guillaume Lample, Lucile Saulnier, Lélio~Renard Lavaud, Marie-Anne Lachaux, Pierre Stock, Teven~Le Scao, Thibaut Lavril, Thomas Wang, Timothée Lacroix, and William~El Sayed.
\newblock Mistral 7b, 2023.
\newblock URL \url{https://arxiv.org/abs/2310.06825}.

\bibitem[Lin et~al.(2022)Lin, Hilton, and Evans]{lin2022truthfulqa}
Stephanie Lin, Jacob Hilton, and Owain Evans.
\newblock Truthfulqa: Measuring how models mimic human falsehoods, 2022.

\bibitem[Lin et~al.(2023)Lin, Tan, Lin, Zheng, Pi, Zhang, Diao, Wang, Zhao, Yao, et~al.]{lin2023speciality}
Yong Lin, Lu~Tan, Hangyu Lin, Zeming Zheng, Renjie Pi, Jipeng Zhang, Shizhe Diao, Haoxiang Wang, Han Zhao, Yuan Yao, et~al.
\newblock Speciality vs generality: An empirical study on catastrophic forgetting in fine-tuning foundation models.
\newblock \emph{arXiv preprint arXiv:2309.06256}, 2023.

\bibitem[Liu et~al.(2024)Liu, Hoang, Zhang, Zhu, Lan, Kokane, Tan, Yao, Liu, Feng, Murthy, Yang, Savarese, Niebles, Wang, Heinecke, and Xiong]{liu2024apigenautomatedpipelinegenerating}
Zuxin Liu, Thai Hoang, Jianguo Zhang, Ming Zhu, Tian Lan, Shirley Kokane, Juntao Tan, Weiran Yao, Zhiwei Liu, Yihao Feng, Rithesh Murthy, Liangwei Yang, Silvio Savarese, Juan~Carlos Niebles, Huan Wang, Shelby Heinecke, and Caiming Xiong.
\newblock Apigen: Automated pipeline for generating verifiable and diverse function-calling datasets, 2024.
\newblock URL \url{https://arxiv.org/abs/2406.18518}.

\bibitem[Luo et~al.(2024)Luo, Yang, Meng, Li, Zhou, and Zhang]{luo2024empiricalstudycatastrophicforgetting}
Yun Luo, Zhen Yang, Fandong Meng, Yafu Li, Jie Zhou, and Yue Zhang.
\newblock An empirical study of catastrophic forgetting in large language models during continual fine-tuning, 2024.
\newblock URL \url{https://arxiv.org/abs/2308.08747}.

\bibitem[McCloskey and Cohen(1989)]{mccloskey1989catastrophic}
Michael McCloskey and Neal~J Cohen.
\newblock Catastrophic interference in connectionist networks: The sequential learning problem.
\newblock In \emph{Psychology of learning and motivation}, volume~24, pages 109--165. Elsevier, 1989.

\bibitem[Nakano et~al.(2022)Nakano, Hilton, Balaji, Wu, Ouyang, Kim, Hesse, Jain, Kosaraju, Saunders, Jiang, Cobbe, Eloundou, Krueger, Button, Knight, Chess, and Schulman]{nakano2022webgptbrowserassistedquestionansweringhuman}
Reiichiro Nakano, Jacob Hilton, Suchir Balaji, Jeff Wu, Long Ouyang, Christina Kim, Christopher Hesse, Shantanu Jain, Vineet Kosaraju, William Saunders, Xu~Jiang, Karl Cobbe, Tyna Eloundou, Gretchen Krueger, Kevin Button, Matthew Knight, Benjamin Chess, and John Schulman.
\newblock Webgpt: Browser-assisted question-answering with human feedback, 2022.
\newblock URL \url{https://arxiv.org/abs/2112.09332}.

\bibitem[{Nam Pham}(2023)]{nam_pham_2023}
{Nam Pham}.
\newblock tiny-textbooks (revision 14de7ba), 2023.
\newblock URL \url{https://huggingface.co/datasets/nampdn-ai/tiny-textbooks}.

\bibitem[Niu et~al.(2024)Niu, Wang, Cheng, Song, and Zhang]{niu2024enhancingdialoguestatetracking}
Cheng Niu, Xingguang Wang, Xuxin Cheng, Juntong Song, and Tong Zhang.
\newblock Enhancing dialogue state tracking models through llm-backed user-agents simulation, 2024.
\newblock URL \url{https://arxiv.org/abs/2405.13037}.

\bibitem[OpenAI(2024)]{openai_api}
OpenAI.
\newblock Openai api.
\newblock \url{https://platform.openai.com/docs/api-reference/introduction}, 2024.

\bibitem[OpenAI et~al.(2024)OpenAI, Achiam, Adler, Agarwal, Ahmad, Akkaya, Aleman, Almeida, Altenschmidt, Altman, Anadkat, Avila, Babuschkin, Balaji, Balcom, Baltescu, Bao, Bavarian, Belgum, Bello, Berdine, Bernadett-Shapiro, Berner, Bogdonoff, Boiko, Boyd, Brakman, Brockman, Brooks, Brundage, Button, Cai, Campbell, Cann, Carey, Carlson, Carmichael, Chan, Chang, Chantzis, Chen, Chen, Chen, Chen, Chen, Chess, Cho, Chu, Chung, Cummings, Currier, Dai, Decareaux, Degry, Deutsch, Deville, Dhar, Dohan, Dowling, Dunning, Ecoffet, Eleti, Eloundou, Farhi, Fedus, Felix, Fishman, Forte, Fulford, Gao, Georges, Gibson, Goel, Gogineni, Goh, Gontijo-Lopes, Gordon, Grafstein, Gray, Greene, Gross, Gu, Guo, Hallacy, Han, Harris, He, Heaton, Heidecke, Hesse, Hickey, Hickey, Hoeschele, Houghton, Hsu, Hu, Hu, Huizinga, Jain, Jain, Jang, Jiang, Jiang, Jin, Jin, Jomoto, Jonn, Jun, Kaftan, Łukasz Kaiser, Kamali, Kanitscheider, Keskar, Khan, Kilpatrick, Kim, Kim, Kim, Kirchner, Kiros, Knight, Kokotajlo, Łukasz Kondraciuk, Kondrich,
  Konstantinidis, Kosic, Krueger, Kuo, Lampe, Lan, Lee, Leike, Leung, Levy, Li, Lim, Lin, Lin, Litwin, Lopez, Lowe, Lue, Makanju, Malfacini, Manning, Markov, Markovski, Martin, Mayer, Mayne, McGrew, McKinney, McLeavey, McMillan, McNeil, Medina, Mehta, Menick, Metz, Mishchenko, Mishkin, Monaco, Morikawa, Mossing, Mu, Murati, Murk, Mély, Nair, Nakano, Nayak, Neelakantan, Ngo, Noh, Ouyang, O'Keefe, Pachocki, Paino, Palermo, Pantuliano, Parascandolo, Parish, Parparita, Passos, Pavlov, Peng, Perelman, de~Avila Belbute~Peres, Petrov, de~Oliveira~Pinto, Michael, Pokorny, Pokrass, Pong, Powell, Power, Power, Proehl, Puri, Radford, Rae, Ramesh, Raymond, Real, Rimbach, Ross, Rotsted, Roussez, Ryder, Saltarelli, Sanders, Santurkar, Sastry, Schmidt, Schnurr, Schulman, Selsam, Sheppard, Sherbakov, Shieh, Shoker, Shyam, Sidor, Sigler, Simens, Sitkin, Slama, Sohl, Sokolowsky, Song, Staudacher, Such, Summers, Sutskever, Tang, Tezak, Thompson, Tillet, Tootoonchian, Tseng, Tuggle, Turley, Tworek, Uribe, Vallone, Vijayvergiya,
  Voss, Wainwright, Wang, Wang, Wang, Ward, Wei, Weinmann, Welihinda, Welinder, Weng, Weng, Wiethoff, Willner, Winter, Wolrich, Wong, Workman, Wu, Wu, Wu, Xiao, Xu, Yoo, Yu, Yuan, Zaremba, Zellers, Zhang, Zhang, Zhao, Zheng, Zhuang, Zhuk, and Zoph]{openai2024gpt4technicalreport}
OpenAI, Josh Achiam, Steven Adler, Sandhini Agarwal, Lama Ahmad, Ilge Akkaya, Florencia~Leoni Aleman, Diogo Almeida, Janko Altenschmidt, Sam Altman, Shyamal Anadkat, Red Avila, Igor Babuschkin, Suchir Balaji, Valerie Balcom, Paul Baltescu, Haiming Bao, Mohammad Bavarian, Jeff Belgum, Irwan Bello, Jake Berdine, Gabriel Bernadett-Shapiro, Christopher Berner, Lenny Bogdonoff, Oleg Boiko, Madelaine Boyd, Anna-Luisa Brakman, Greg Brockman, Tim Brooks, Miles Brundage, Kevin Button, Trevor Cai, Rosie Campbell, Andrew Cann, Brittany Carey, Chelsea Carlson, Rory Carmichael, Brooke Chan, Che Chang, Fotis Chantzis, Derek Chen, Sully Chen, Ruby Chen, Jason Chen, Mark Chen, Ben Chess, Chester Cho, Casey Chu, Hyung~Won Chung, Dave Cummings, Jeremiah Currier, Yunxing Dai, Cory Decareaux, Thomas Degry, Noah Deutsch, Damien Deville, Arka Dhar, David Dohan, Steve Dowling, Sheila Dunning, Adrien Ecoffet, Atty Eleti, Tyna Eloundou, David Farhi, Liam Fedus, Niko Felix, Simón~Posada Fishman, Juston Forte, Isabella Fulford, Leo
  Gao, Elie Georges, Christian Gibson, Vik Goel, Tarun Gogineni, Gabriel Goh, Rapha Gontijo-Lopes, Jonathan Gordon, Morgan Grafstein, Scott Gray, Ryan Greene, Joshua Gross, Shixiang~Shane Gu, Yufei Guo, Chris Hallacy, Jesse Han, Jeff Harris, Yuchen He, Mike Heaton, Johannes Heidecke, Chris Hesse, Alan Hickey, Wade Hickey, Peter Hoeschele, Brandon Houghton, Kenny Hsu, Shengli Hu, Xin Hu, Joost Huizinga, Shantanu Jain, Shawn Jain, Joanne Jang, Angela Jiang, Roger Jiang, Haozhun Jin, Denny Jin, Shino Jomoto, Billie Jonn, Heewoo Jun, Tomer Kaftan, Łukasz Kaiser, Ali Kamali, Ingmar Kanitscheider, Nitish~Shirish Keskar, Tabarak Khan, Logan Kilpatrick, Jong~Wook Kim, Christina Kim, Yongjik Kim, Jan~Hendrik Kirchner, Jamie Kiros, Matt Knight, Daniel Kokotajlo, Łukasz Kondraciuk, Andrew Kondrich, Aris Konstantinidis, Kyle Kosic, Gretchen Krueger, Vishal Kuo, Michael Lampe, Ikai Lan, Teddy Lee, Jan Leike, Jade Leung, Daniel Levy, Chak~Ming Li, Rachel Lim, Molly Lin, Stephanie Lin, Mateusz Litwin, Theresa Lopez, Ryan
  Lowe, Patricia Lue, Anna Makanju, Kim Malfacini, Sam Manning, Todor Markov, Yaniv Markovski, Bianca Martin, Katie Mayer, Andrew Mayne, Bob McGrew, Scott~Mayer McKinney, Christine McLeavey, Paul McMillan, Jake McNeil, David Medina, Aalok Mehta, Jacob Menick, Luke Metz, Andrey Mishchenko, Pamela Mishkin, Vinnie Monaco, Evan Morikawa, Daniel Mossing, Tong Mu, Mira Murati, Oleg Murk, David Mély, Ashvin Nair, Reiichiro Nakano, Rajeev Nayak, Arvind Neelakantan, Richard Ngo, Hyeonwoo Noh, Long Ouyang, Cullen O'Keefe, Jakub Pachocki, Alex Paino, Joe Palermo, Ashley Pantuliano, Giambattista Parascandolo, Joel Parish, Emy Parparita, Alex Passos, Mikhail Pavlov, Andrew Peng, Adam Perelman, Filipe de~Avila Belbute~Peres, Michael Petrov, Henrique~Ponde de~Oliveira~Pinto, Michael, Pokorny, Michelle Pokrass, Vitchyr~H. Pong, Tolly Powell, Alethea Power, Boris Power, Elizabeth Proehl, Raul Puri, Alec Radford, Jack Rae, Aditya Ramesh, Cameron Raymond, Francis Real, Kendra Rimbach, Carl Ross, Bob Rotsted, Henri Roussez,
  Nick Ryder, Mario Saltarelli, Ted Sanders, Shibani Santurkar, Girish Sastry, Heather Schmidt, David Schnurr, John Schulman, Daniel Selsam, Kyla Sheppard, Toki Sherbakov, Jessica Shieh, Sarah Shoker, Pranav Shyam, Szymon Sidor, Eric Sigler, Maddie Simens, Jordan Sitkin, Katarina Slama, Ian Sohl, Benjamin Sokolowsky, Yang Song, Natalie Staudacher, Felipe~Petroski Such, Natalie Summers, Ilya Sutskever, Jie Tang, Nikolas Tezak, Madeleine~B. Thompson, Phil Tillet, Amin Tootoonchian, Elizabeth Tseng, Preston Tuggle, Nick Turley, Jerry Tworek, Juan Felipe~Cerón Uribe, Andrea Vallone, Arun Vijayvergiya, Chelsea Voss, Carroll Wainwright, Justin~Jay Wang, Alvin Wang, Ben Wang, Jonathan Ward, Jason Wei, CJ~Weinmann, Akila Welihinda, Peter Welinder, Jiayi Weng, Lilian Weng, Matt Wiethoff, Dave Willner, Clemens Winter, Samuel Wolrich, Hannah Wong, Lauren Workman, Sherwin Wu, Jeff Wu, Michael Wu, Kai Xiao, Tao Xu, Sarah Yoo, Kevin Yu, Qiming Yuan, Wojciech Zaremba, Rowan Zellers, Chong Zhang, Marvin Zhang, Shengjia
  Zhao, Tianhao Zheng, Juntang Zhuang, William Zhuk, and Barret Zoph.
\newblock Gpt-4 technical report, 2024.
\newblock URL \url{https://arxiv.org/abs/2303.08774}.

\bibitem[Osika(2023)]{Osika_gpt-engineer_2023}
Anton Osika.
\newblock {gpt-engineer}, April 2023.
\newblock URL \url{https://github.com/gpt-engineer-org/gpt-engineer}.

\bibitem[Parisi et~al.(2022)Parisi, Zhao, and Fiedel]{parisi2022talmtoolaugmentedlanguage}
Aaron Parisi, Yao Zhao, and Noah Fiedel.
\newblock Talm: Tool augmented language models, 2022.
\newblock URL \url{https://arxiv.org/abs/2205.12255}.

\bibitem[Ratcliff(1990)]{ratcliff1990connectionist}
Roger Ratcliff.
\newblock Connectionist models of recognition memory: constraints imposed by learning and forgetting functions.
\newblock \emph{Psychological review}, 97\penalty0 (2):\penalty0 285, 1990.

\bibitem[Sakaguchi et~al.(2019)Sakaguchi, Bras, Bhagavatula, and Choi]{DBLP:journals/corr/abs-1907-10641}
Keisuke Sakaguchi, Ronan~Le Bras, Chandra Bhagavatula, and Yejin Choi.
\newblock {WINOGRANDE:} an adversarial winograd schema challenge at scale, 2019.

\bibitem[Setlur et~al.(2024)Setlur, Garg, Geng, Garg, Smith, and Kumar]{setlur2024rl}
Amrith Setlur, Saurabh Garg, Xinyang Geng, Naman Garg, Virginia Smith, and Aviral Kumar.
\newblock Rl on incorrect synthetic data scales the efficiency of llm math reasoning by eight-fold.
\newblock \emph{arXiv preprint arXiv:2406.14532}, 2024.

\bibitem[Shi et~al.(2024)Shi, Xu, Wang, Qin, Wang, Wang, Wang, Ebrahimi, and Wang]{shi2024continuallearninglargelanguage}
Haizhou Shi, Zihao Xu, Hengyi Wang, Weiyi Qin, Wenyuan Wang, Yibin Wang, Zifeng Wang, Sayna Ebrahimi, and Hao Wang.
\newblock Continual learning of large language models: A comprehensive survey, 2024.
\newblock URL \url{https://arxiv.org/abs/2404.16789}.

\bibitem[{Significant Gravitas}()]{Significant_Gravitas_AutoGPT}
{Significant Gravitas}.
\newblock {AutoGPT}.
\newblock URL \url{https://github.com/Significant-Gravitas/AutoGPT}.

\bibitem[Tang et~al.(2023)Tang, Deng, Lin, Han, Liang, Cao, and Sun]{tang2023toolalpacageneralizedtoollearning}
Qiaoyu Tang, Ziliang Deng, Hongyu Lin, Xianpei Han, Qiao Liang, Boxi Cao, and Le~Sun.
\newblock Toolalpaca: Generalized tool learning for language models with 3000 simulated cases, 2023.
\newblock URL \url{https://arxiv.org/abs/2306.05301}.

\bibitem[Team et~al.(2024)Team, Mesnard, Hardin, Dadashi, Bhupatiraju, Pathak, Sifre, Rivière, Kale, Love, Tafti, Hussenot, Sessa, Chowdhery, Roberts, Barua, Botev, Castro-Ros, Slone, Héliou, Tacchetti, Bulanova, Paterson, Tsai, Shahriari, Lan, Choquette-Choo, Crepy, Cer, Ippolito, Reid, Buchatskaya, Ni, Noland, Yan, Tucker, Muraru, Rozhdestvenskiy, Michalewski, Tenney, Grishchenko, Austin, Keeling, Labanowski, Lespiau, Stanway, Brennan, Chen, Ferret, Chiu, Mao-Jones, Lee, Yu, Millican, Sjoesund, Lee, Dixon, Reid, Mikuła, Wirth, Sharman, Chinaev, Thain, Bachem, Chang, Wahltinez, Bailey, Michel, Yotov, Chaabouni, Comanescu, Jana, Anil, McIlroy, Liu, Mullins, Smith, Borgeaud, Girgin, Douglas, Pandya, Shakeri, De, Klimenko, Hennigan, Feinberg, Stokowiec, hui Chen, Ahmed, Gong, Warkentin, Peran, Giang, Farabet, Vinyals, Dean, Kavukcuoglu, Hassabis, Ghahramani, Eck, Barral, Pereira, Collins, Joulin, Fiedel, Senter, Andreev, and Kenealy]{gemmateam2024gemmaopenmodelsbased}
Gemma Team, Thomas Mesnard, Cassidy Hardin, Robert Dadashi, Surya Bhupatiraju, Shreya Pathak, Laurent Sifre, Morgane Rivière, Mihir~Sanjay Kale, Juliette Love, Pouya Tafti, Léonard Hussenot, Pier~Giuseppe Sessa, Aakanksha Chowdhery, Adam Roberts, Aditya Barua, Alex Botev, Alex Castro-Ros, Ambrose Slone, Amélie Héliou, Andrea Tacchetti, Anna Bulanova, Antonia Paterson, Beth Tsai, Bobak Shahriari, Charline~Le Lan, Christopher~A. Choquette-Choo, Clément Crepy, Daniel Cer, Daphne Ippolito, David Reid, Elena Buchatskaya, Eric Ni, Eric Noland, Geng Yan, George Tucker, George-Christian Muraru, Grigory Rozhdestvenskiy, Henryk Michalewski, Ian Tenney, Ivan Grishchenko, Jacob Austin, James Keeling, Jane Labanowski, Jean-Baptiste Lespiau, Jeff Stanway, Jenny Brennan, Jeremy Chen, Johan Ferret, Justin Chiu, Justin Mao-Jones, Katherine Lee, Kathy Yu, Katie Millican, Lars~Lowe Sjoesund, Lisa Lee, Lucas Dixon, Machel Reid, Maciej Mikuła, Mateo Wirth, Michael Sharman, Nikolai Chinaev, Nithum Thain, Olivier Bachem,
  Oscar Chang, Oscar Wahltinez, Paige Bailey, Paul Michel, Petko Yotov, Rahma Chaabouni, Ramona Comanescu, Reena Jana, Rohan Anil, Ross McIlroy, Ruibo Liu, Ryan Mullins, Samuel~L Smith, Sebastian Borgeaud, Sertan Girgin, Sholto Douglas, Shree Pandya, Siamak Shakeri, Soham De, Ted Klimenko, Tom Hennigan, Vlad Feinberg, Wojciech Stokowiec, Yu~hui Chen, Zafarali Ahmed, Zhitao Gong, Tris Warkentin, Ludovic Peran, Minh Giang, Clément Farabet, Oriol Vinyals, Jeff Dean, Koray Kavukcuoglu, Demis Hassabis, Zoubin Ghahramani, Douglas Eck, Joelle Barral, Fernando Pereira, Eli Collins, Armand Joulin, Noah Fiedel, Evan Senter, Alek Andreev, and Kathleen Kenealy.
\newblock Gemma: Open models based on gemini research and technology, 2024.
\newblock URL \url{https://arxiv.org/abs/2403.08295}.

\bibitem[TensorOpera(2024)]{foxllm}
TensorOpera.
\newblock Tensoropera unveils fox foundation model: A pioneering small language model (slm) for cloud and edge.
\newblock https://blog.tensoropera.ai/tensoropera-unveils-fox-foundation-model-a-pioneering-open-source-slm-leading-the-way-against-tech-giants/, 2024.

\bibitem[Weng(2023)]{weng2023agent}
Lilian Weng.
\newblock Llm-powered autonomous agents.
\newblock \emph{lilianweng.github.io}, Jun 2023.
\newblock URL \url{https://lilianweng.github.io/posts/2023-06-23-agent/}.

\bibitem[Yan et~al.(2024)Yan, Mao, Ji, Zhang, Patil, Stoica, and Gonzalez]{berkeley-function-calling-leaderboard}
Fanjia Yan, Huanzhi Mao, Charlie Cheng-Jie Ji, Tianjun Zhang, Shishir~G. Patil, Ion Stoica, and Joseph~E. Gonzalez.
\newblock Berkeley function calling leaderboard.
\newblock 2024.

\bibitem[Zelikman et~al.(2022)Zelikman, Wu, Mu, and Goodman]{zelikman2022starbootstrappingreasoningreasoning}
Eric Zelikman, Yuhuai Wu, Jesse Mu, and Noah~D. Goodman.
\newblock Star: Bootstrapping reasoning with reasoning, 2022.
\newblock URL \url{https://arxiv.org/abs/2203.14465}.

\bibitem[Zellers et~al.(2019)Zellers, Holtzman, Bisk, Farhadi, and Choi]{zellers2019hellaswag}
Rowan Zellers, Ari Holtzman, Yonatan Bisk, Ali Farhadi, and Yejin Choi.
\newblock Hellaswag: Can a machine really finish your sentence?, 2019.

\end{thebibliography}
